\documentclass[10pt,twocolumn,letterpaper]{article}

\usepackage{wacv}              %

\usepackage[most]{tcolorbox}
\usepackage[ruled,vlined,linesnumbered]{algorithm2e}
\usepackage{algpseudocode}
\usepackage{eqparbox}
\usepackage{comment}
\usepackage{multirow}
\usepackage{makecell}
\usepackage{wrapfig}

\newcommand{\methodname}{TEA-Prompt}

\makeatletter
\DeclareRobustCommand\onedot{\futurelet\@let@token\@onedot}
\def\@onedot{\ifx\@let@token.\else.\null\fi\xspace}
\makeatother

\def\etal{\emph{et al}\onedot}

\algnewcommand\LeftComment[2]{%
\hspace{#1\algindent}$\triangleright$ \eqparbox{COMMENT}{\textbf{\textit{#2}}} \hfill %
}

\definecolor{wacvblue}{rgb}{0.21,0.49,0.74}
\usepackage[pagebackref,breaklinks,colorlinks,allcolors=wacvblue]{hyperref}

\def\wacvPaperID{*****} %
\def\confName{WACV}
\def\confYear{2027}

\title{Anchor Behavior, Not Parameters: Rehearsal-Free Continual Learning with a Single Shared Prompt}

\author{
Duy-Cuong Bui\textsuperscript{1}\quad
Trung-Anh Dang\textsuperscript{2}\quad
Vincent Nguyen\textsuperscript{2}\quad
Duc V. Le\textsuperscript{3}\quad
Ngoc-Son Vu\textsuperscript{1,4}\\[3pt]
\textsuperscript{1}ETIS - CY Cergy Paris Université, ENSEA, CNRS UMR 8051, France\\
\textsuperscript{2}LIFO UR 4022, Université d’Orléans, INSA CVL, France\quad
\textsuperscript{3}University of Twente, Netherlands\\
\textsuperscript{4}LIST3N - Université de Technologie de Troyes, France\\
{\tt\small duy-cuong.bui@ensea.fr, \{trung-anh.dang, vincent.nguyen\}@univ-orleans.fr,}\\
{\tt\small v.d.le@utwente.nl, son.vu@utt.fr}
}

\begin{document}
\maketitle
\begin{abstract}
Prompt-based continual learning adapts a frozen pre-trained vision transformer by tuning a small set of prompt tokens. Recently, one-prompt methods which share one single prompt across all tasks make this paradigm maximally parameter efficient. The open question is how a shared prompt can keep the knowledge of old classes while being updated for new ones. Existing methods protect old knowledge in \emph{parameter space}, where they freeze task-specific prompts, expand new components, or sparsely gate which tokens receive gradients. However, this method sacrifices plasticity, yet the drift it still permits in shared tokens can corrupt the prompt's function on past classes. What must be preserved is the prompt's \emph{behavior}. We propose \methodname{}, which couples (i) a \emph{token evaluation} module that assigns each prompt token a continuous, input-conditional relevance score injected into the attention logits, softly specializing tokens across tasks without discarding any of them, and (ii) an \emph{attention-behavior anchoring} loss that constrains each token's attention output around class-specific query regions using only compact Gaussian query statistics without stored old images, weighted by the token's past relevance. Anchoring the attention value allows the parameters to continue updating freely, as long as their interaction remains unchanged. \methodname{} achieves state-of-the-art accuracy with markedly lower forgetting on CIFAR-100, ImageNet-R, and CUB-200 with only 0.38M parameters. Code is publicy available at https://github.com/cuonga1cvp/TEA-Prompt.git
\end{abstract}

\section{Introduction}
\label{sec:intro}

Continual learning (CL) requires a model to learn new knowledge from a stream of tasks without catastrophically forgetting \cite{mccloskey_1989, robins1995_catastrophic_fr, dangSAMixCalibratedAccurate2026a, anh_wacv_2025}. With large pre-trained vision transformers \cite{dosovitskiy2020image}, prompt-based CL has become a leading rehearsal-free paradigm where the backbone is frozen to protect its general representations, and the adaptation is limited to a small set of learnable tokens \cite{wang2022learning, wang2022dualprompt, smith2023coda}.

Within this paradigm, most methods allocate task-specific prompts or expand prompt components as tasks arrive \cite{wang2022dualprompt, smith2023coda, wang2023hierarchical, hong2025rainbowprompt, jiao2024vector}, so prompt memory and selection cost grow linearly with the number of tasks. The opposite design, a \emph{single prompt shared across all tasks} \cite{huang2024ovor, le2025one}, keeps parameters and compute constant regardless of the task count. It also highlights the main difficulty: When a shared prompt adapts to new tasks, what prevents it from forgetting old knowledge?

Examining existing methods reveals a common pattern: they act in \emph{parameter space}. \emph{(i) Isolation and freezing:} task-specific prompts are frozen once their task ends \cite{wang2022dualprompt, wang2023spromptslearningpretrainedtransformers, wang2023hierarchical}, and expansion-based methods freeze previous components while training new ones \cite{smith2023coda}. \emph{(ii) Selection sparsity:} only retrieved prompts are updated in pool-based methods \cite{wang2022learning}, and sparse Top-$K$ gating over a decomposed shared prompt lets only the selected components receive gradients, complemented by a prototype loss on prompt keys \cite{le2025one}. Across these two families, the implicit assumption is the same: old knowledge is safe if and only if old \emph{parameters} stay (nearly) where they were.

We argue this assumption is flawed on both sides. Firstly, it is \emph{too strong}: Freezing and expansion tie stability to capacity, and sparse gating leaves unselected tokens idle. All paying in plasticity or parameters. Secondly, it is \emph{too weak}: What an old class actually needs is that the prompt keeps producing the same \emph{attention contribution} for its inputs. In classical CL, this is precisely the argument for regularizing in \emph{function space} rather than weight space \cite{benjamin2019function, titsias2020functional}, and attention-level preservation has proven effective for fully trainable ViTs \cite{pelosin2022towards}. Prompt-based CL, however, still lacks a function-space preservation mechanism. Even though, with a frozen backbone, the only source of attention drift is the prompt itself, making its behavior both necessary and sufficient object to protect.

In this paper, we propose \methodname{}, which protects the shared prompt's \emph{behavior} rather than its \emph{parameters}, through two coupled components. First, a \textbf{token evaluation module} treats the shared prompt as a set of key-value token pairs and assigns each token a continuous, input-conditional relevance score, computed from the correlation between the input query and the token's key and passed through a sigmoid-based evaluation function. Injected as a penalty into the attention logits, the score lets relevant tokens dominate adaptation for the current input while attenuating the rest, inducing a soft, task-conditional specialization of tokens. Second, an \textbf{attention-behavior anchoring loss} stores compact class-wise Gaussian statistics of reduced queries. When learning new tasks, it samples queries from these distributions and penalizes deviations of each token's attention output from its previous behavior. Crucially, the two components are coupled: the deviation of each token is weighted by its past relevance score, so preservation concentrates exactly on the tokens that mattered for old classes, while low-relevance tokens are free to adapt aggressively.

The main contributions of our work are as follows:
\begin{itemize}
    \item We recast anti-forgetting in shared-prompt CL as \emph{behavior preservation} instead of \emph{parameter preservation}, and instantiate it with a class-conditional, data-free attention anchoring loss using compact Gaussian query statistics and the previous prompt tokens.
    \item We introduce a token evaluation module that assigns continuous, input-conditional relevance to each prompt token, softly specializing tokens across tasks to reduce cross-task interference; its stored past scores direct the anchoring loss toward the tokens where preservation matters.
    \item Extensive experiments on CIFAR-100, ImageNet-R, and CUB-200, including self-supervised backbones, show state-of-the-art accuracy and the lowest forgetting among shared-prompt methods at efficient parameter and compute cost.
\end{itemize}

\section{Related Work}
\label{sec:related}

\subsection{Continual learning}
CL aims to incrementally acquire new knowledge while preserving previously learned information. The primary challenge in CL is catastrophic forgetting, where optimizing for new tasks leads to a severe decline in performance on earlier ones. To address this issue, traditional solutions are categorized into three main approaches: (1) \textit{regularization-based} methods mitigate forgetting by introducing penalty terms that protect the weights critical to prior tasks \cite{kirkpatrick2017overcoming, aljundi2018memory, benzing2022unifying, li2017learning}; (2) \textit{replay-based} methods maintain performance by either storing a subset of old data or using generative techniques to reconstruct past information \cite{rebuffi2017icarl, aljundi2019gradient, chaudhry2021using}; (3) \textit{architecture-based} methods dynamically adjust the network structure or isolate specific parameter clusters to accommodate each new task \cite{mallya2018piggyback, serra2018overcoming}. While these methods have made significant strides, the emergence of pre-trained models \cite{dosovitskiy2020image} trained on massive datasets has triggered a paradigm shift. Recent trends have moved away from resource-intensive retraining toward parameter-efficient approaches (e.g., Prompt Tuning \cite{lester2021powerscaleparameterefficientprompt}, LoRA \cite{hu2022lora}) and replay-free methods \cite{wang2022learning, wang2022dualprompt, smith2023coda}, harnessing powerful pre-trained generalization capabilities for more efficient learning. These methods also differ in where to preserve: weight-space penalties \cite{kirkpatrick2017overcoming, friedemann_si_2017} versus function-space constraints \cite{li2017learning, benjamin2019function, titsias2020functional}. Our work brings the function-space perspective to prompt-based CL.

\subsection{Prompt-based continual learning}
L2P \cite{wang2022learning} pioneered prompt-based CL with a shared pool from which task-specific prompts are retrieved by key-query matching; DualPrompt \cite{wang2022dualprompt} decomposes prompts into general and task-specific components attached at different layers; CODA-Prompt \cite{smith2023coda} replaces discrete retrieval with an attention-weighted composition over per-task components, enabling end-to-end optimization; VQ-Prompt \cite{jiao2024vector} quantizes input-conditioned prompts into discrete codes, and further variants refine prompt evolution and consistency \cite{hong2025rainbowprompt, gao2024consistent, wang2023hierarchical}. These methods perform strongly, but their prompt parameters and selection cost grow with the number of tasks. A second line keeps a single shared prompt: OVOR \cite{huang2024ovor} pairs it with virtual-outlier regularization of the classifier, while SMoPE \cite{le2025one} views prefix tuning through a mixture-of-experts perspective. 

Across both lines, forgetting is combated at the parameter level: task-specific prompts are frozen after their task \cite{wang2022dualprompt, wang2023spromptslearningpretrainedtransformers, wang2023hierarchical} or previous components are frozen under expansion \cite{smith2023coda}; or hard selection restricts which tokens receive gradients at all \cite{wang2022learning, le2025one}. None of these directly constrains what the prompt does on previously seen classes. On the contrary, our method protects the behavior of token that correlated to the old task and provides a smooth token evaluation to maintain the ability of learning new tasks.

\subsection{Preserving behavior versus parameters}
Function-level preservation has a long history in CL: LwF distills output logits \cite{li2017learning}, PodNet distills pooled intermediate features \cite{douillard2020podnet}, and continual distillation has recently been explored in prompt-based CL across teacher-student models \cite{zhang2024continual}. These objectives operate at the task level and require a stored teacher copy evaluated on current-task data. Closer to our target, Pelosin \etal \cite{pelosin2022towards} distill pooled pre-softmax attention maps between consecutive, fully trainable ViTs for exemplar-free CL, establishing attention as an effective lever against forgetting. Our loss differs in scope and machinery: with a frozen backbone, we anchor only the prompt tokens' attention outputs by storing class-based Gaussian query statistics, requiring neither old data nor a network copy, with the constraint weighted by each token's past relevance.

\section{Methodology}
\label{sec:method}

\begin{figure*}[t]
    \centering
    \includegraphics[width=1\linewidth]{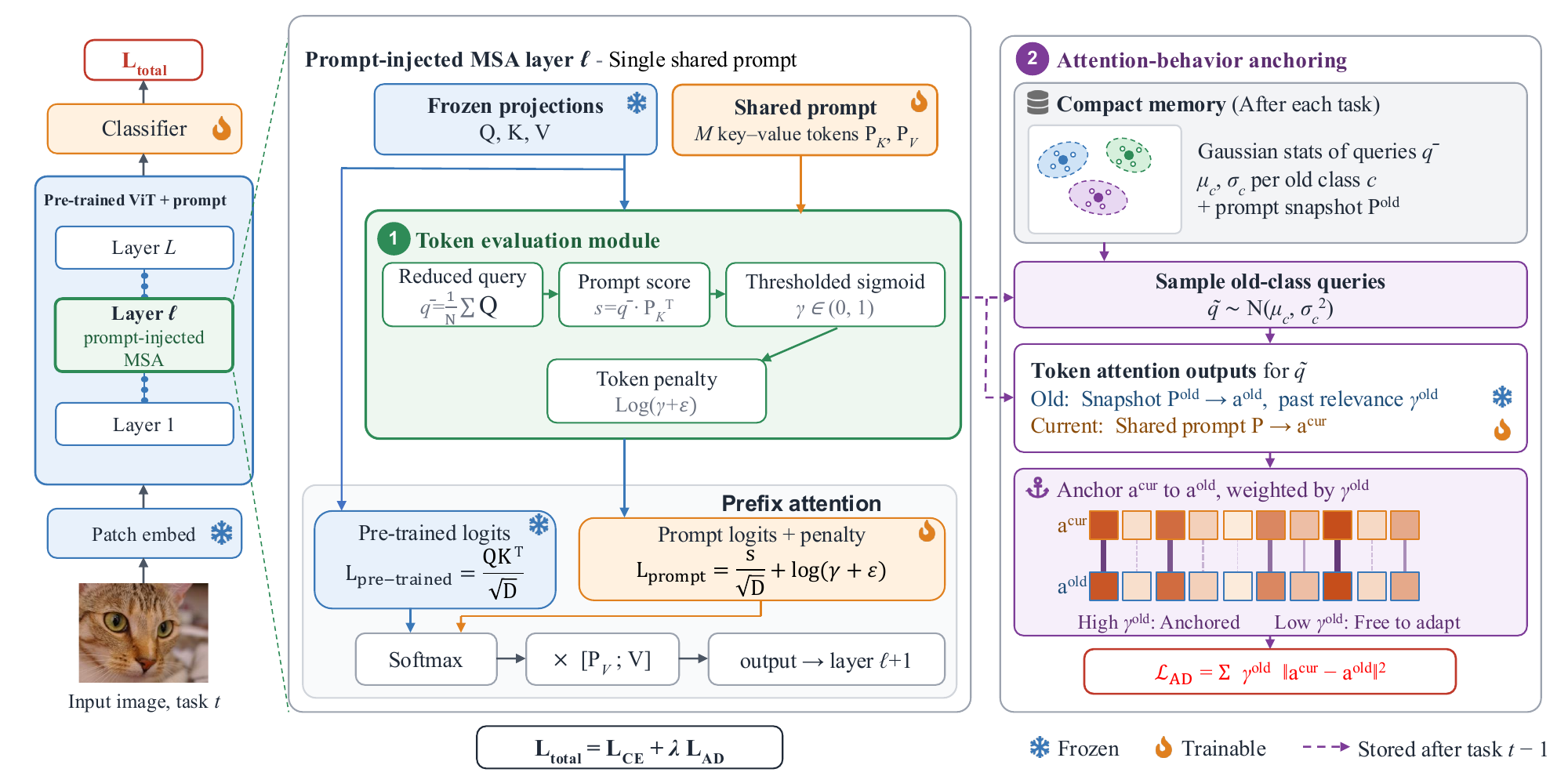}
    \caption{\textbf{Overview of \methodname{}.} \textbf{Left:} Inside
    each MSA layer of the frozen ViT, prompt tokens
    $(\mathbf{P}_K,\mathbf{P}_V)$ are prepended to the frozen keys and
    values. The reduced query scores every prompt token
    (Eq.~\eqref{eq:scores}); a thresholded sigmoid turns the scores into
    evaluation weights (Eq.~\eqref{eq:gate}) whose logarithm shifts the
    prompt attention logits $L_{\mathrm{prompt}}$ (Eq.~\eqref{eq:logits}),
    modulating each token's attention mass before the joint softmax with
    the input-token logits $L_{\mathrm{pre\text{-}trained}}$
    (Eq.~\eqref{eq:attn}). \textbf{Right:} At the end of task $t{-}1$ the
    prompt is snapshotted and class-conditional Gaussian query statistics
    are stored (Eq.~\eqref{eq:stats}); while learning task $t$, queries
    sampled from these statistics (Eq.~\eqref{eq:sample}) probe the prompt
    attention of the frozen snapshot ($A^{(t-1)}$) and of the current
    prompt ($A^{(t)}$), and $\mathcal{L}_{\mathrm{AD}}$
    (Eq.~\eqref{eq:lad}) anchors the current behavior to the old one,
    weighted by each token's past relevance.}
    \label{fig:method}
\end{figure*}

\methodname{} adapts a frozen vision transformer with a \emph{single}
prompt shared by all tasks and protects it with two components: a
\emph{token evaluation module} that decides, per instance, how strongly
each prompt token participates in attention
(Sec.~\ref{sec:tea}), and an \emph{attention-behavior anchoring loss}
that preserves what the prompt tokens \emph{do} for past classes rather
than what their parameters \emph{are} (Sec.~\ref{sec:anchor}).
Section~\ref{sec:training} assembles the training objective and
procedure. Figure~\ref{fig:method} gives an overview.

\subsection{Preliminaries}
\label{sec:prelim}

\paragraph{Continual learning setting.}
We consider class-incremental learning over a sequence of $T$ tasks.
Task $t$ arrives with a dataset
$\mathcal{D}_t=\{(x_i,y_i)\}_{i=1}^{N_t}$ whose label set
$\mathcal{C}_t$ is disjoint from all other tasks; task identity is
unknown at inference. Following~\cite{wang2022learning,wang2022dualprompt}, we adopt the
rehearsal-free scenario: no samples from past tasks are stored.

\paragraph{Prefix tuning on a frozen ViT.}
Let $f_\theta$ be a pre-trained ViT~\cite{dosovitskiy2020image} with frozen parameters
$\theta$. Each transformer block contains a multi-head self-attention
(MSA) layer with $H$ heads; $C$ denotes the embedding dimension and
$D=C/H$ the head dimension. Given the token sequence
$\mathbf{X}\equiv\mathbf{X}^{(\ell)}\in\mathbb{R}^{N\times C}$
entering block $\ell$ ($N$ tokens: one class token and $N{-}1$ patch
tokens), head $h$ computes the frozen projections
\begin{equation}
\mathbf{Q}=\mathbf{X}\mathbf{W}_Q,\quad
\mathbf{K}=\mathbf{X}\mathbf{W}_K,\quad
\mathbf{V}=\mathbf{X}\mathbf{W}_V
\in\mathbb{R}^{N\times D}.
\label{eq:qkv}
\end{equation}
Prompt-based methods adapt $f_\theta$ by prepending learnable prompts
to the frozen keys and values (\emph{prefix
tuning}~\cite{Li2021PrefixTuningOC}). We use a single prompt shared across all
tasks~\cite{huang2024ovor,le2025one}: in each of the first $L_p$ blocks, the prompt
consists of $M$ key-value \emph{prompt tokens} per head,
$\{(\mathbf{p}_{K,m},\mathbf{p}_{V,m})\}_{m=1}^{M}$, stacked as
$\mathbf{P}_K,\mathbf{P}_V\in\mathbb{R}^{M\times D}$. Only the prompt
and the classifier are trained. All equations are written per head,
per injected layer, and per instance; the superscript $(\ell,h)$ and
the batch axis are omitted when clear from context, parameters are
broadcast along the batch axis, and $\top$ denotes the
transpose of a two-dimensional matrix.

\subsection{Token evaluation module}
\label{sec:tea}

A shared prompt must serve every task. However, not every prompt token is relevant to every input. \methodname{} scores each prompt
token based on the current instance and converts the scores into
continuous evaluation weights that modulate the token's
impact inside the MSA (Fig.~\ref{fig:method}, left).

\paragraph{Token scores.}
Prompt tokens score $\mathbf{s}$ are evaluated once per instance and head using the
\emph{reduced query} $\bar{\mathbf{q}}$, which is the average of the token queries:
\begin{equation}
\bar{\mathbf{q}}=\frac{1}{N}\sum_{i=1}^{N}\mathbf{Q}_{i,:}\in\mathbb{R}^{D},
\qquad
\mathbf{s}=\bar{\mathbf{q}}\,\mathbf{P}_K^{\top}\in\mathbb{R}^{M}.
\label{eq:scores}
\end{equation}
Instance-level evaluation has two advantages over per-token scoring:
It produces one consistent set of weights per instance, avoiding the
token-level competition that causes information loss in per-token
sparse selection schemes~\cite{puigcerver2024sparse}, and it reduces the scoring
cost per head from $NM$ to $M$ inner products.

\paragraph{Evaluation weights.}
The raw scores are mapped to weights by a temperature-controlled
sigmoid $\sigma(\cdot)$ centred at an instance-adaptive threshold,
applied element-wise:
\begin{equation}
\boldsymbol{\gamma}
=\sigma\!\left(\frac{\mathbf{s}-\tau(\mathbf{s})}{\tau_{\mathrm{soft}}}\right)
\in(0,1)^{M},
\label{eq:gate}
\end{equation}
where the threshold $\tau(\mathbf{s})$ is the $K$-th largest entry of
$\mathbf{s}$ and $\tau_{\mathrm{soft}}$ is a temperature.
Three properties make Eq.~\eqref{eq:gate} a useful replacement for
discrete selection. \emph{(i) It is a continuous relaxation of
Top-$K$:} as $\tau_{\mathrm{soft}}\!\to\!0$,
$\gamma_m\!\to\!\mathbf{1}[s_m>\tau]$ and hard Top-$K$ selection is
recovered (the boundary token receives weight $\tfrac12$), while
larger $\tau_{\mathrm{soft}}$ interpolates smoothly toward plain dense
prefix tuning. \emph{(ii) Every
prompt token receives gradient:}
$\partial\gamma_m/\partial s_m=\gamma_m(1-\gamma_m)/\tau_{\mathrm{soft}}>0$,
bounded by $1/(4\tau_{\mathrm{soft}})$, so tokens scoring just below
the threshold are attenuated rather than eliminated and can recover in
later tasks. \emph{(iii) Weights are independent across tokens:}
unlike softmax-based weighting, the sigmoid imposes no sum-to-one
competition among prompt tokens, a choice that avoids unnecessary competition between
components~\cite{nguyen2024sigmoid}, so several prompt tokens can be
fully active for the same instance.

\paragraph{Injection into the attention logits.}
The evaluation weights modulate each prompt token multiplicatively
inside the MSA. We form one row of prompt logits per instance and
head,
\begin{equation}
\mathbf{L}_{\mathrm{prompt}}
=\frac{\mathbf{s}}{\sqrt{D}}
+\log\!\big(\boldsymbol{\gamma}+\varepsilon\big)
\in\mathbb{R}^{1\times M},
\label{eq:logits}
\end{equation}
with a small $\varepsilon>0$ for numerical stability, and concatenate
it with the frozen token--token logits before a joint softmax:
\begin{equation}
\label{eq:attn}
\begin{aligned}
\mathbf{A}=\big[\,\mathbf{1}_N\,\mathbf{L}_{\mathrm{prompt}}||\;
\mathbf{L}_{\mathrm{pre\text{-}trained}}\big]\in\mathbb{R}^{N\times(M+N)},\\
\mathbf{O}=\mathrm{softmax}(\mathbf{A})
\begin{bmatrix}\mathbf{P}_V\\ \mathbf{V}\end{bmatrix}\in\mathbb{R}^{N\times D},
\end{aligned}
\end{equation}
where $\mathbf{1}_N\in\mathbb{R}^{N\times 1}$ broadcasts the single
row of prompt logits to all $N$ queries and $\|$ denotes column-wise
concatenation. Because the softmax is taken jointly over the $M{+}N$
columns, adding $\log\gamma_m$ multiplies the unnormalized attention
mass of prompt token $m$ by $(\gamma_m+\varepsilon)$. The evaluation
weight acts as an input-conditioned prior over prompt tokens: $\gamma_m\!\to\!0$ removes token $m$ from the attention distribution
and thereby suppresses its value $\mathbf{p}_{V,m}$, while
$\gamma_m\!\to\!1$ leaves it untouched. Although the prompt logits are
shared across the $N$ queries, the post-softmax prompt weights remain
query-dependent.
The forward pass of
Eqs.~\eqref{eq:scores}-\eqref{eq:attn} is identical during training
and inference.

\subsection{Anchoring prompt behavior}
\label{sec:anchor}

Sequential training drifts the attention behavior of the shared prompt, which is the main cause of forgetting. Prior methods constrain prompt {parameters}: Weights are regularized toward, or grown alongside,
their past values. \methodname{} instead anchors prompt
{behavior}: What the prompt tokens contribute to attention when
queried from the representation space occupied by past
classes (Fig.~\ref{fig:method}, right). The anchor is rehearsal-free, only stores compact per-class query statistics and one snapshot of the prompt.

\paragraph{Class-conditional query statistics.}
We model the class-conditional distribution of the reduced query with a
diagonal Gaussian. After finishing task $t$, we run one inference pass
over $\mathcal{D}_t$ and record, for every injected layer, head, and
class $c\in\mathcal{C}_t$,
\begin{equation}
\boldsymbol{\mu}_c=\frac{1}{N_c}\!\!\sum_{n:\,y_n=c}\!\!\bar{\mathbf{q}}_n,
\qquad
\boldsymbol{\sigma}_c^{2}
=\frac{1}{N_c}\!\!\sum_{n:\,y_n=c}\!\!\bar{\mathbf{q}}_n^{2}
-\boldsymbol{\mu}_c^{2},
\label{eq:stats}
\end{equation}
where $N_c$ is the number of training samples of class $c$, squares
are element-wise, and variances are clamped from below for stability.
This Gaussian assumption is mild for the \emph{reduced} query:
averaging over $N$ token queries concentrates each class's query
distribution toward a compact, unimodal shape, and per-dimension Gaussian statistics of deep features
are an established data-free memory
mechanism~\cite{wang2023hierarchical}.
We further keep the statistics \emph{per class} rather than per task:
distillation applied at the task level constrains only an aggregate of
past behavior and fails to capture the fine-grained, class-specific
attention behavior encoded in the prompt tokens, sacrificing
discriminative information for previously seen
classes~\cite{douillard2020podnet,pelosin2022towards,zhang2024continual}.
Class-conditional sampling instead probes the prompts' behavior
separately in the query region of every seen class. %
Since class sets are disjoint, statistics of earlier classes are left
unchanged. The prompt is then snapshotted,
$\mathbf{P}^{\mathrm{old}}\!\leftarrow\!\mathbf{P}$.

\begin{algorithm}[t]
\caption{Training procedure of \methodname{}}
\label{alg:training}
\KwIn{frozen ViT $f_\theta$; classifier $\phi$; prompt
$\mathbf{P}=\{\mathbf{P}_K,\mathbf{P}_V\}$ initialized
$\mathcal{U}[0,1)$; tasks $\{\mathcal{D}_t\}_{t=1}^{T}$; epochs $E$;
anchor weight $\lambda$; samples $r$; rank $K$; temperature
$\tau_{\mathrm{soft}}$; TAP epochs $E_{\mathrm{TAP}}$}
\KwOut{$\mathbf{P},{\phi}$}
$\mathcal{C}_{\mathrm{seen}}\leftarrow\emptyset$\;
\For{$t=1,\dots,T$}{
  \If{$t=1$}{
    train $\mathbf{P},\phi$ on $\mathcal{D}_1$ with
    $\mathcal{L}_{\mathrm{CE}}$ for $E/2$ epochs
    \tcp*[r]{stabilization}
    reset optimizer\;
  }
  \For{$e=1,\dots,E$}{
    \ForEach{$(x,y)\in\mathcal{D}_t$}{
      forward pass with token evaluation,
      Eqs.~\eqref{eq:scores}--\eqref{eq:attn};
      predict $\hat{y}=\phi\big(f_\theta(x;\mathbf{P})\big)$\;
      \eIf{$t>1$}{
        draw $\{\tilde{\mathbf{q}}_{c,j}\}$ for all
        $c\in\mathcal{C}_{\mathrm{seen}}$, Eq.~\eqref{eq:sample};
        compute $\mathcal{L}_{\mathrm{AD}}$, Eq.~\eqref{eq:lad}\;
      }{
        $\mathcal{L}_{\mathrm{AD}}\leftarrow 0$\;
      }
      update $\mathbf{P},\phi$ with
      $\mathcal{L}_{\mathrm{CE}}+\lambda\,\mathcal{L}_{\mathrm{AD}}$
      \tcp*[r]{Eq.~\eqref{eq:total}}
    }
  }
  estimate $\boldsymbol{\mu}_c,\boldsymbol{\sigma}_c^{2}$ of
  $\bar{\mathbf{q}}$ over $\mathcal{D}_t$ for each
  $c\in\mathcal{C}_t$, Eq.~\eqref{eq:stats}
  \tcp*[r]{one inference pass}
  $\mathbf{P}^{\mathrm{old}}\leftarrow\mathbf{P}$;\quad
  $\mathcal{C}_{\mathrm{seen}}\leftarrow
  \mathcal{C}_{\mathrm{seen}}\cup\mathcal{C}_t$\;
  fine-tune $\phi$ for $E_{\mathrm{TAP}}$ epochs with TAP
  pseudo-features~\cite{wang2023hierarchical}\;
}
\Return{$\mathbf{P},\phi$}
\end{algorithm}

\paragraph{Behavioral anchor.}
While learning any subsequent task, at \emph{every} optimization step
we draw $r$ fresh queries per seen class from the stored Gaussians,
\begin{equation}
\tilde{\mathbf{q}}_{c,j}
=\boldsymbol{\mu}_c+\boldsymbol{\sigma}_c\odot\boldsymbol{\epsilon}_j,
\qquad
\boldsymbol{\epsilon}_j\sim\mathcal{N}(\mathbf{0},\mathbf{I}),
\quad j=1,\dots,r,
\label{eq:sample}
\end{equation}
and compare, token by token, the attention read-out that the old and
the current prompt produce for these queries. With scaled scores
$\tilde{\mathbf{s}}^{\mathrm{old}}
=\tilde{\mathbf{q}}\,(\mathbf{P}_K^{\mathrm{old}})^{\top}/\sqrt{D}$
and
$\tilde{\mathbf{s}}^{\mathrm{cur}}
=\tilde{\mathbf{q}}\,\mathbf{P}_K^{\top}/\sqrt{D}$
(class, sample, layer, and head indices omitted), the read-out of
prompt token $m$ is its softmax-weighted value contribution,
\begin{equation}
\mathbf{a}^{\mathrm{old},(m)}
=\big[\mathrm{softmax}\big(\tilde{\mathbf{s}}^{\mathrm{old}}\big)\big]_m
\,\mathbf{p}^{\mathrm{old}}_{V,m}\in\mathbb{R}^{D},
\label{eq:readout}
\end{equation}
and analogously $\mathbf{a}^{\mathrm{cur},(m)}$ from
$\tilde{\mathbf{s}}^{\mathrm{cur}}$ and $\mathbf{p}_{V,m}$ of the
current prompt. The softmax is restricted to the $M$ prompt tokens: the anchor
targets the prompts' own attention behavior, not a full model output.
Each token's error is weighted by its past relevance to the
sampled query,
$\gamma^{\mathrm{old}}_{m}(\tilde{\mathbf{q}})$, so that only tokens that
mattered for past classes are constrained. The anchoring loss averages
the weighted errors and sums over injected layers and heads:
\begin{equation}
\mathcal{L}_{\mathrm{AD}}
=\frac{1}{Z}
\sum_{\ell,h}\,\sum_{m,c,j}
\gamma^{\mathrm{old}}_{m}\big(\tilde{\mathbf{q}}_{c,j}\big)\,
\big\|\mathbf{a}^{\mathrm{cur},(m)}_{c,j}
-\mathbf{a}^{\mathrm{old},(m)}_{c,j}\big\|_2^2\,,
\label{eq:lad}
\end{equation}
with normalizer $Z=M\,|\mathcal{C}_{\mathrm{seen}}|\,r\,D$, where the
sums run over injected layers $\ell\le L_p$, heads $h\le H$, prompt
tokens $m\le M$, seen classes $c\in\mathcal{C}_{\mathrm{seen}}$, and
samples $j\le r$; the dependence of $\tilde{\mathbf{q}}_{c,j}$ and of
the read-outs on $(\ell,h)$ is left implicit.
The anchor stores $2\,L_p H\,|\mathcal{C}_{\mathrm{seen}}|\,D$
floating-point values (about $3.7$\,MB in single precision for
$|\mathcal{C}_{\mathrm{seen}}|{=}100$ with our configuration) plus one
prompt snapshot; its computational cost grows linearly with
$|\mathcal{C}_{\mathrm{seen}}|\,r$ and is incurred only during
training. Unlike attention-distillation approaches that align full
attention maps between a frozen model copy and the current model on
current-task images~\cite{pelosin2022towards}, Eq.~\eqref{eq:lad} is
class-conditional, data-free, and restricted to the prompt.

\subsection{Training objective and procedure}
\label{sec:training}

\paragraph{Objective.}
The model is optimized with the classification loss and the behavioral
anchor:
\begin{equation}
\mathcal{L}_{\mathrm{total}}
=\mathcal{L}_{\mathrm{CE}}+\lambda\,\mathcal{L}_{\mathrm{AD}},
\label{eq:total}
\end{equation}
where $\mathcal{L}_{\mathrm{CE}}$ is the cross-entropy over the
classes of the current task, logits of previously seen classes are masked while learning
task $t$, and the classifier is re-balanced afterwards, and $\mathcal{L}_{\mathrm{AD}}=0$ for the first task. The
frozen backbone receives no gradient; only the prompt
$\{\mathbf{P}_K,\mathbf{P}_V\}$, initialized i.i.d.\ uniformly on
$[0,1)$, and the classifier are updated.

\paragraph{Training-only components.}
Three components are used exclusively during training: (i) the
statistics pass of Eq.~\eqref{eq:stats}, one additional inference pass
over $\mathcal{D}_t$ per task; (ii) the anchoring loss with its
per-step sampling, Eqs.~\eqref{eq:sample}-\eqref{eq:lad}; and (iii)
the classifier re-balancing step, and the token evaluation module behaves identically at
training and test time.

\paragraph{First-task stabilization.}
For the first task we prepend a stabilization phase of $E/2$ epochs
followed by an optimizer reset before the main $E$-epoch schedule;
warm-starting the shared prompt in this way reduces early instability
of the evaluation weights.

\paragraph{Classifier re-balancing.}
Class-incremental training biases a shared linear classifier toward
recently added classes. After each task we therefore re-balance the
classifier with task-adaptive prediction (TAP) ~\cite{wang2023hierarchical,jiao2024vector}:
Class-wise Gaussian statistics of the prompted features provide
pseudo-features of all seen classes, on which only the classifier is
fine-tuned for $E_{\mathrm{TAP}}$ epochs. The full procedure is
summarized in Algorithm~\ref{alg:training}.

\section{Experiments}
\subsection{Datasets and Implementation Details}

\noindent \textbf{Datasets.} To evaluate \methodname{} performance, we perform comprehensive image classification in the class-incremental learning settings using three primary benchmarks: CIFAR-100 \cite{krizhevsky2009learning}, ImageNet-R \cite{hendrycks2021many}, and CUB-200 \cite{welinder2010caltech}. CIFAR-100 (100 classes) serves as a standard benchmark for prompt-based CL to evaluate generalization. ImageNet-R is selected for its inherent difficulty, containing 200 classes with diverse artistic styles and high intra-class variance. Finally, we utilize the fine-grained CUB-200 dataset of 200 bird species to test the model's ability to distinguish subtle inter-class differences. Across all experiments, each dataset is partitioned into 10 tasks with disjoint classes: 20 classes/task for ImageNet-R, 10 classes/task for CIFAR-100, and 20 classes/task for CUB-200.

\noindent \textbf{Evaluation metrics.} 
To evaluate performance, we use two standard metrics: Average Accuracy (Avg Acc) and Forgetting \cite{chaudhry_2018}. Avg Acc measures the model's performance across all tasks after completing the training sequence, while Forgetting quantifies the performance drop on previously learned tasks. Let $R_{t,i}$ denote the accuracy on task $i$ after training on task $t$. The Avg Acc $\mathcal{A}_T$ after the final task $T$ is defined as:

\begin{equation}
    \mathcal{A}_T = \frac{1}{T}\sum_{k=1}^{T} R_{T, k}
\end{equation}

Similarly, the Forgetting metric $\mathcal{F}$ is defined as the mean performance drop from each task's initial accuracy to its final state:

\begin{equation}
    \mathcal{F} = \frac{1}{T-1} \sum_{i=1}^{T-1}(R_{i,i}-R_{T,i})
\end{equation}

\noindent \textbf{Compared methods.} We evaluate the performance of \methodname{} against a comprehensive suite of methods, including non-prompt-based methods such as LAE \cite{gao2023lae}, InfLoRA \cite{liang2024inflora}, and EASE \cite{zhou2024ease}, as well as established prompt-based architectures: L2P \cite{wang2022learning}, DualPrompt \cite{wang2022dualprompt}, CODA-Prompt \cite{smith2023coda}, OVOR-Deep \cite{huang2024ovor}, ESN \cite{wang2023isolation}, RainbowPrompt \cite{hong2025rainbowprompt}, CPrompt~\cite{gao2024consistent}, VQ-Prompt \cite{jiao2024vector}, and SMoPE \cite{le2025one}. The results are reproduced based on the original setting in these works.

\noindent \textbf{Implementation details.} All experiments are executed on a NVIDIA RTX PRO 6000 Blackwell. Following \cite{hong2025rainbowprompt, wang2022learning}, we primarily use the pre-trained ViT-B/16 on ImageNet-1K \cite{russakovsky2015imagenet} and ImageNet-21K \cite{ridnik2021imagenet} as our backbone. \methodname{} is optimized using the AdamW optimizer \cite{loshchilov2017decoupled} with a cosine learning rate decay schedule to stabilize training. For the training configurations, we employ a batch size of 64 for ImageNet-R, while a larger batch size of 128 is utilized for both CIFAR-100 and CUB-200. Besides, the classifier in CL tends to favor recently added classes, leading to biased predictions \cite{hou_learning_2019, 9009019}. To mitigate this effect, we apply a simple \text{task adaptive prediction} (TAP) technique inspired by \cite{jiao2024vector, wang2023hierarchical}. The model predictions are adjusted using feature statistics from previously learned classes, which helps rebalance old and new categories and stabilizes prompt learning throughout training.

On CIFAR-100, \methodname{} uses $M{=}25$ key--value prompt-token pairs per attention head (a shared prompt of length $50$), evaluated with rank $K{=}5$ and temperature $\tau_{\mathrm{soft}}{=}0.2$ and injected into the first $L_p{=}6$ transformer blocks; prompt keys and values are initialized i.i.d.\ uniformly on $[0,1)$, and $\varepsilon{=}10^{-6}$. The anchoring loss uses $r{=}10$ sampled queries per seen class with weight $\lambda{=}60$, and the classifier re-balancing step runs for $E_{\mathrm{TAP}}{=}50$ epochs. The settings for other benchmarks are mentioned in the supplementary material. The complete training procedure is given in Algorithm~\ref{alg:training}.

\subsection{Experimental Results} 
\label{sec:result}

\begin{table*}[t]
\centering
\caption{Performance comparison on CIFAR-100, ImageNet-R, and CUB-200 datasets. Params counts the learnable parameters on ImageNet-R with 10 tasks. ( "-" means Not Available)}
\label{tab:mainresults}
\small %
\setlength{\tabcolsep}{4.5pt}
\begin{tabular}{lccccccc}
\toprule
\multirow{2}{*}{\textbf{Method}} & \multirow{2}{*}{\textbf{Params (M)}} & \multicolumn{2}{c}{\textbf{CIFAR-100}} & \multicolumn{2}{c}{\textbf{ImageNet-R}} & \multicolumn{2}{c}{\textbf{CUB-200}} \\
\cmidrule(lr){3-4} \cmidrule(lr){5-6} \cmidrule(lr){7-8}
& & Avg Acc($\uparrow$) & Forgetting($\downarrow$) & Avg Acc($\uparrow$) & Forgetting($\downarrow$) & Avg Acc($\uparrow$) & Forgetting($\downarrow$) \\
\midrule
Joint training & -- & 91.38 & -- & 82.06 & -- & 88.00 & -- \\
\midrule
LAE \cite{gao2023lae}        & 0.24 & 86.09 $\pm$ 0.55 & 6.22 $\pm$ 0.17 & 73.88 $\pm$ 0.43 & 5.08 $\pm$ 0.33 & -- & -- \\ %
InfLoRA \cite{liang2024inflora}       & 3.84 & 86.91 $\pm$ 0.34 & 4.22 $\pm$ 0.37 & 76.13 $\pm$ 0.44 & 5.22 $\pm$ 0.48 & -- & -- \\ %
EASE \cite{zhou2024ease}        & 11.90 & 87.28 $\pm$ 0.71 & 4.81 $\pm$ 0.42 & 75.82 $\pm$ 0.54 & 5.83 $\pm$ 0.37 & -- & -- \\ %
\midrule
L2P \cite{wang2022learning}           & 0.24 & 83.86 $\pm$ 0.28 & 7.35 $\pm$ 0.38& 61.57 $\pm$ 0.66 & 9.73 $\pm$ 0.47  & 71.22 $\pm$ 0.54  & 10.68 $\pm$ 0.61 \\ %
DualPrompt \cite{wang2022dualprompt}    & 1.10 & 86.51 $\pm$ 0.33 & 5.16 $\pm$ 0.09 &  68.13 $\pm$ 0.49  & 4.68 $\pm$ 0.20  &71.55 $\pm$ 0.73  & 10.21 $\pm$ 0.43 \\
CODA-Prompt \cite{smith2023coda}   & 3.99& 86.92 $\pm$ 0.77& 5.10 $\pm$ 0.43&   73.21 $\pm$ 0.28  & 5.69 $\pm$ 0.28 & 73.25 $\pm$ 0.71 & 11.02 $\pm$ 0.88 \\
OVOR-Deep \cite{huang2024ovor}     & 0.26 & 85.99 $\pm$ 0.89 & 6.42 $\pm$ 2.03 & 76.11 $\pm$ 0.21 & 7.16 $\pm$ 0.34 & 78.11 $\pm$ 0.47 & 7.95 $\pm$ 0.77 \\
ESN \cite{wang2023isolation}         & 0.92 & 86.42 $\pm$ 0.80 & 6.08 $\pm$ 0.48 & 75.11 $\pm$ 0.36  & 5.68 $\pm$ 0.77  & 71.20 $\pm$ 0.54 & 9.82 $\pm$ 0.43 \\ %
RainbowPrompt \cite{hong2025rainbowprompt} & 6.60 & 87.22 $\pm$ 0.53 & 4.81 $\pm$ 0.41 & 76.65 $\pm$ 0.56 & 4.57 $\pm$ 0.33 & 70.43 $\pm$ 1.23 & 10.53 $\pm$ 0.91 \\ %
CPrompt \cite{gao2024consistent}      & 1.08 &  87.82 $\pm$ 0.21 & 5.06 $\pm$ 0.50& 77.14 $\pm$ 0.11   & 5.97 $\pm$ 0.68  & 77.09  $\pm$ 0.64& 10.27 $\pm$ 0.54 \\ %
VQ-Prompt \cite{jiao2024vector}  & 0.50 & 88.73 $\pm$ 0.27 & 4.97 $\pm$ 0.24 & 76.96 $\pm$ 0.22 & 6.09 $\pm$ 0.54 & 86.72 $\pm$ 0.94 & 4.53 $\pm$ 0.32 \\
SMoPE \cite{le2025one} & 0.38 & 88.86 $\pm$ 0.39 & 4.57 $\pm$ 0.45 & 76.70 $\pm$ 0.67 & 5.71 $\pm$ 0.72 & 85.73 $\pm$ 0.87 & 5.61 $\pm$ 0.64 \\
\midrule
\methodname{} (Ours)  & 0.38 & \textbf{90.59} $\pm$ 0.31 & \textbf{2.68} $\pm$ 0.25 & \textbf{77.40} $\pm$ 0.42 & \textbf{4.19} $\pm$ 0.39 & \textbf{86.84} $\pm$ 0.23 & \textbf{3.30} $\pm$ 0.35 \\
\bottomrule
\end{tabular}
\end{table*}

\begin{table*}[!t]
    \begin{center}
        \caption{Average accuracy on CIFAR-100 and ImageNet-R with self-supervised pre-trained models iBOT-1K and DINO-1K. Best results are in \textbf{bold}.} 
        \label{tab:results_pretrained_v2}
        \small %
        \setlength{\tabcolsep}{6mm}
        \begin{tabular}{llcc}
            \toprule 
            \textbf{Pre-trained model} & \textbf{Method} & \textbf{CIFAR-100} & \textbf{ImageNet-R} \\
            \cmidrule(lr){1-1} \cmidrule(lr){2-2} \cmidrule(lr){3-3} \cmidrule(lr){4-4} 
            \multirow{7}{*}{iBOT-1K}
            & L2P  & $75.57 \pm 0.41$ & $60.97 \pm 0.26$ \\
            & DualPrompt  & $76.63 \pm 0.05$ & $61.51 \pm 1.05$ \\
            & CODA-Prompt  & $76.42 \pm 0.40$ & $66.56 \pm 0.68$ \\
            & RainbowPrompt  & $76.20 \pm 0.34$ & $65.82 \pm 0.52$ \\
            & VQ-Prompt  &  $ 75.80 \pm 0.34$ & $71.68 \pm 0.72$ \\
            & SMoPE  & $ 76.10 \pm 0.34$ & $72.17 \pm 0.52$ \\
            & {\methodname{} (Ours)} & $\mathbf{76.73} \pm 0.57$ & $\mathbf{72.61} \pm 0.32$ \\
            
            \midrule
            
            \multirow{7}{*}{DINO-1K}
            & L2P  & $70.65 \pm 0.57$ & $57.40 \pm 0.23$ \\
            & DualPrompt  & $74.90 \pm 0.21$ & $58.57 \pm 0.45$ \\
            & CODA-Prompt  & $77.50 \pm 0.64$ & $63.15 \pm 0.39$ \\
            & RainbowPrompt  & $74.77 \pm 0.54$ & $62.04 \pm 0.43$ \\
            & VQ-Prompt  & $ 78.32 \pm 0.31$ & $68.42 \pm 0.28$ \\
            & SMoPE  & $ 78.64 \pm 0.44$ & $68.61 \pm 0.52$ \\
            & {\methodname{} (Ours)} & $\mathbf{79.10} \pm 0.38$ & $\mathbf{68.94} \pm 0.33$ \\
            
            \bottomrule
        \end{tabular}
    \end{center}
\end{table*}

\paragraph{Main results.}
Results on three benchmarks 
are summarized in Table \ref{tab:mainresults}.
\begin{itemize}
    \item \textbf{CIFAR-100:} On this standard benchmark, \methodname{} achieves notable performance with an average accuracy of 90.59\%, which lies a mere 0.79\% below the upper bound. When compared with SMoPE, which relies on sparse Top-$K$ selection over a shared prompt structure, \methodname{} yields a 1.73\% improvement in accuracy while reducing the forgetting rate from 4.57\% down to 2.68\% with the same number of parameters. This clear margin demonstrates that our continuous token evaluation can address the rigidity of discrete token selection
    by exploiting complementary information from lower-ranked tokens. Furthermore, \methodname{} outperforms recent weighted prompt composition methods such as VQ-Prompt (88.73\%) and CODA-Prompt (86.92\%), suggesting the superior efficacy of our rehearsal-free loss  in mitigating forgetting without requiring access to past training samples.
    
    \item \textbf{ImageNet-R:} On the challenging ImageNet-R dataset, characterized by severe distribution shifts and high intra-class variance across diverse artistic styles, \methodname{} achieves the state-of-the-art (SOTA) performance with an average accuracy (Avg Acc) of 77.40\% and a forgetting rate of 4.19\%. Furthermore, when compared to OVOR, a method utilizing a shared prompt architecture, our method reduces the forgetting rate by approximately 3\% but remains a small number of trainable parameters. This further underscores the effectiveness of TEA-Prompt in mitigating forgetting within challenging scenarios.

    \item \textbf{CUB-200:} On the CUB-200 fine-grained classification dataset, while OVOR struggled with classification, achieving an accuracy of only 78.11\%, \methodname{} boosted accuracy to 86.84\%, despite sharing the same shared-prompt architecture. Meanwhile, although SMoPE achieved good accuracy, it exhibited a relatively high forgetting rate of 5.61\%. We attribute this to the fact that while their anti-forgetting mechanism employs a prototype loss function to preserve old prefix keys, the prefix values remain susceptible to drift, resulting in an incomplete distillation process. In contrast, by preserving prior knowledge through the anchoring of attention behaviors for old tokens, %
    \methodname{} achieved better resistance to forgetting, with a rate of only 3.30\%. This result suggests that our method can handle forgetting under different circumstances, from standard benchmark such as CIFAR-100 to fine-grained dataset such as CUB-200.
\end{itemize}

\paragraph{Results on different self-supervised pre-trained models.}
As shown in Table \ref{tab:results_pretrained_v2}, our proposed \methodname{} consistently outperforms existing prompt-based continual learning methods across different self-supervised pre-trained models \cite{zhou2022ibotimagebertpretraining, caron2021emergingpropertiesselfsupervisedvision} and benchmark datasets. This demonstrates that \methodname{} can effectively leverage self-supervised representations while improving the stability and generalization ability of prompt learning.

\subsection{Ablation study}

\label{subsec:ablation}

\begin{table}[t]
\centering
\caption{Ablating different components of our method on CIFAR-100.}
\label{tab:ablation_study}
\small %
\setlength{\tabcolsep}{5pt}
\begin{tabular}{lcc}
\toprule
\textbf{Training Strategy} & \textbf{Avg Acc} ($\uparrow$) & \textbf{Forgetting} ($\downarrow$) \\
\midrule
Baseline & 83.44 $\pm$ 0.19 & 6.48 $\pm$ 0.42 \\
+ Token Evaluation & 86.73 $\pm$ 0.45 & 5.32 $\pm$ 0.36 \\
+ Task-Adaptive Prediction & 88.73 $\pm$ 0.29 & 3.74 $\pm$ 0.43 \\
+ First-Task Stabilization & 89.25 $\pm$ 0.21 & 3.68 $\pm$ 0.38 \\

+ Anchoring Loss $\mathcal{L}_{\mathrm{AD}}$ & 90.59 $\pm$ 0.31 & 2.68 $\pm$ 0.25 \\
\bottomrule
\end{tabular}
 \end{table}

To evaluate the individual contribution of each component within our proposed framework, we conduct a comprehensive ablation study by incrementally integrating them into the training pipeline. As summarized in Table~\ref{tab:ablation_study}, we analyze the impact of each module on both Avg Acc and Forgetting metrics.

The baseline configuration utilizes a single shared prompt across all tasks without any token weighting or preservation constraints, yielding an Avg Acc of 83.44\% and a Forgetting rate of 6.48\%. This result highlights the vulnerability of static shared parameters to catastrophic forgetting when adapted sequentially.

\begin{table}[t]
    \centering
    \caption{Estimated anchor overhead (FP32, ViT-B/16, $L_p{=}6$,
    $H{=}12$, $D{=}64$, $M{=}25$, $r{=}10$). Per-step cost is for
    learning the \emph{final} task, probing the
    $|\mathcal{C}_{\mathrm{seen}}|$ previously seen classes; memory is
    the complete anchor state after training (statistics of all
    classes plus the snapshot $\mathbf{P}^{\mathrm{old}}$).}
    \label{tab:anchor_overhead}
    \small
    \setlength{\tabcolsep}{4pt}
    \renewcommand{\arraystretch}{1.1}
    \begin{tabular}{lccc}
        \toprule
        Dataset & $|\mathcal{C}_{\mathrm{seen}}|$
        & $\mathcal{L}_{\mathrm{AD}}$/step & Memory \\
        \midrule
        CIFAR-100            & 90  & 0.94 GFLOPs & 4.61 MB \\
        ImageNet-R / CUB-200 & 180 & 1.88 GFLOPs & 8.29 MB \\
        \bottomrule
    \end{tabular}
\end{table}

\begin{table}[t]
\centering
\caption{Ablation study for the interaction between the token evaluation module and anchoring loss on CIFAR-100, 10 tasks.}
\label{tab:anchor_weighting}
\small
\setlength{\tabcolsep}{6pt}
\begin{tabular}{lcc}
\toprule
Anchor weighting & Avg Acc ($\uparrow$) & Forgetting ($\downarrow$) \\
\midrule
Inverse relevance ($1{-}\gamma^{\mathrm{old}}$) & 89.05 $\pm$ 0.20& 4.55 $\pm$ 0.35\\
Uniform ($\gamma^{\mathrm{old}}\!\equiv\!1$)     & 90.10 $\pm$ 0.35  & 3.22 $\pm$ 0.27  \\
Past relevance (ours)      & 90.59 $\pm$ 0.31 & 2.68 $\pm$ 0.25 \\
\bottomrule
\end{tabular}
\end{table}

\begin{itemize}
    \item \textbf{Token evaluation:} Introducing the token evaluation module decomposes the shared prompt into continuous weighted prompt tokens. This module significantly boosts the performance, pushing the Avg Acc to 86.73\% while reducing Forgetting to 5.32\%. 
    This improvement validates the benefit of our continuous token-evaluation module over rigid parameter constraints.

    \item \textbf{Anchoring loss:} Finally, integrating the complete
framework with the proposed anchoring loss ($\mathcal{L}_{\mathrm{AD}}$)
achieves 
the highest Avg Acc of 90.59\%
and lower the Forgetting rate from 3.68\% to 2.68\%. %
This final improvement shows that our rehearsal-free preservation objective is effective. It preserves the attention distributions of old classes without using any historical raw data.
The per-step Monte-Carlo sampling behind this gain is cheap: drawing $r{=}10$ queries per seen class and evaluating
Eqs.~\eqref{eq:sample}--\eqref{eq:lad} over all injected layers and
heads costs $\approx\!0.94$~GFLOPs, 4.61MB (for CIFAR-100 dataset) and $\approx\!1.88$~GFLOPs, 8.29MB (for ImageNet-R/CUB-200 datasets). As the loss is training-only, inference cost is unchanged.
    \item \textbf{Interaction between the token evaluation module and anchoring loss:} Table~\ref{tab:anchor_weighting} illustrates the interaction between the two components by varying weights based on historical relevance ($\gamma^{\mathrm{old}}$) in Equation~\eqref{eq:lad}. Removing the importance weights from the anchor (uniform scheme, $\gamma^{\mathrm{old}}\!\equiv\!1$) degrades performance across both metrics ($90.10$ vs. $90.59$ in Average Accuracy; $3.22$ vs. $2.68$ in Forgetting). In this case, the anchor retains tokens that offer no related information about old classes, thereby prevents the updating of those tokens for new tasks. Consequently, applying weights based on historical relevance is not only a performance optimization for mitigating forgetting but also facilitates learning of new tasks. Furthermore, reversing the weighting scheme ($1{-}\gamma^{\mathrm{old}}$) yields the poorest results ($89.05$ / $4.55$). This is a consequence of tokens associated with the old task being allowed to drift, while unrelated tokens remain anchored. In conclusion, these results reaffirm that the token importance module and the anchoring loss function are tightly integrated components rather than independent add-ons.

\end{itemize}

\section{Conclusion}
In this paper, we recast anti-forgetting for shared-prompt continual learning as a problem of preserving the prompt's \emph{behavior} rather than its parameters. \methodname{} couples a token evaluation module, assigning a continuous, input-conditional relevance score to every prompt token, with an attention-behavior anchoring loss that constrains token attention outputs around class-specific Gaussian query statistics, weighted by past relevance so that preservation concentrates where it mattered. Extensive experiments on standard class-incremental benchmarks show that \methodname{} consistently outperforms strong prompt-based baselines across datasets and pre-trained backbones while keeping parameters and compute independent of the task count. These results validate relevance-guided behavior anchoring as a scalable, rehearsal-free route to continual learning with pre-trained models.

\section{Acknowledgement}
\label{sec: Acknowledgement}
This work was supported by the France Excellence scholarship from the French Ministry for Europe and Foreign Affairs.
 
{
    \small
    \bibliographystyle{ieeenat_fullname}
    \bibliography{ref}
}

\appendix
\twocolumn[{\centering{\Large\bf Supplementary Material\par}\vspace{1.5em}}]

\begin{table*}[t]
    \caption{Hyperparameters used in \methodname{}.}
    \centering
    \small
    \setlength{\tabcolsep}{4pt}
    \renewcommand{\arraystretch}{1.1}
        \begin{tabular}{clccc}
            \toprule
            \textbf{Hyperparameter} & \textbf{Description} & \textbf{CIFAR-100} & \textbf{ImageNet-R} & \textbf{CUB-200} \\
            \midrule
            $E$  & Number of training epochs & 50 & 50 & 50 \\
            $\tilde{\eta}$ & Learning rate & 0.005 & 0.0025 & 0.0025 \\
            $bsz$ & Batch size & 128 & 64 & 128 \\
            $n$ & Token prompt length & 50 & 50 & 50 \\
            $\tau(\mathbf{s})$ & Threshold for the soft weighting function & 5-th highest score & 5-th highest score & 5-th highest score \\
            $\tau_{\mathrm{soft}}$ & Temperature for the soft weighting function  & 0.2 & 0.5 & 0.2 \\
            $\varepsilon$ & Small constant for numerical stability  & 1e-6 & 1e-6 & 1e-6 \\
            $\lambda$ & Coefficient of $\mathcal{L}_{\mathrm{AD}}$  & 60 & 1e-2 & 100 \\
            $r$ & Number of sampled queries per class in $\mathcal{L}_{\mathrm{AD}}$ & 10 & 10 & 20 \\
            Prompt position & Positions in MSA layer to inject prompt & [0, 1, 2, 3, 4, 5] & [0, 1, 2, 3, 4, 5] & [0, 1, 2, 3, 4, 5] \\
            \bottomrule
        \end{tabular}
    \label{tab:search-space-hyperparameters}
\end{table*}

\section{Additional training details}
\label{supp:training_details}

Table~\ref{tab:search-space-hyperparameters} lists all hyperparameters,
split into two groups. The first group is \emph{not searched}: The
backbone, prompt length $n{=}50$, injection into the first six MSA blocks, learning rate, batch size,
number of epochs, and the classifier re-balancing schedule
($E_{\mathrm{TAP}}{=}50$) are inherited from the shared-prompt setting
of SMoPE~\cite{le2025one} and kept identical to the baselines
reproduced under the same backbone. The second group is searched per
dataset: the temperature $\tau_{\mathrm{soft}}$, the evaluation rank
$K$ defining the threshold $\tau(\mathbf{s})$, the anchor weight
$\lambda$, and the number of sampled queries $r$. Because the four
values interact only weakly, we use a coordinate-wise search rather
than a full grid: Starting from ($\tau_{\mathrm{soft}}{=}0.2$, $K{=}5$,
$r{=}10$, $\lambda{=}60$), obtained from a coarse preliminary sweep on
CIFAR-100, we vary one hyperparameter at a time over
$\tau_{\mathrm{soft}}\!\in\!\{0.01,0.05,0.1,0.2,0.4,0.8,1.6,3.2,6.4\}$,
$K\!\in\!\{1,2,3,5,10,25\}$, $r\!\in\!\{1,2,4,\dots,20\}$, and
$\lambda\!\in\!\{0.1,1,5,10,50,60,70,80,100,150,200\}$ (34 distinct
single-seed runs, Table~\ref{tab:sensitivity}), and retain the value
maximizing final average accuracy. Starting from
the CIFAR-100 optimum, we therefore transfer $\lambda$ to the other
benchmarks by keeping the effective ratio
$\lambda\,\mathcal{L}_{\mathrm{AD}}/\mathcal{L}_{\mathrm{CE}}$
approximately constant.

The anchor weight varies widely across datasets ($\lambda{=}60$, $0.01$,
and $100$ on CIFAR-100, ImageNet-R, and CUB-200) because $\lambda$
multiplies a quantity whose scale is dataset-dependent:
$\mathcal{L}_{\mathrm{AD}}$ is the mean squared deviation of the prompt
attention outputs at sampled queries, so the wider
the class-conditional query distributions, the farther the samples fall
from the class centroids, the larger the raw loss, and the smaller the
$\lambda$ that yields the same effective anchor strength.  ImageNet-R, whose 200 classes each span
paintings, cartoons, and sketches, has the largest per-class spread and hence
the smallest $\lambda$, whereas the fine-grained CUB-200 has the
tightest distributions so it gets the largest $\lambda$.

\begin{table*}[t]
\centering
\caption{Sensitivity of \methodname{} to $\tau_{\mathrm{soft}}$, $K$ in $\tau(\mathbf{s})$, $r$,
and $\lambda$ on CIFAR-100 (10 tasks, single seed). Each sweep varies
one hyperparameter with the other three fixed at the shared optimum
($\tau_{\mathrm{soft}}{=}0.2$, $K=5$, $r{=}10$, $\lambda{=}60$, marked
$^{\dagger}$), which corresponds to the main-paper configuration.}
\label{tab:sensitivity}
\small
\setlength{\tabcolsep}{4pt}
\renewcommand{\arraystretch}{1.1}
\begin{tabular}{lccccccccc}
\toprule
$\tau_{\mathrm{soft}}$ & 0.01 & 0.05 & 0.1 & 0.2$^{\dagger}$ & 0.4 & 0.8 & 1.6 & 3.2 & 6.4 \\
\midrule
Avg Acc ($\uparrow$)      & 89.78 & 90.02 & 90.36 & 90.59 & 90.04 & 89.36 & 89.20 & 90.23 & 89.53 \\
Forgetting ($\downarrow$) & 3.25  & 3.00  & 2.90  & 2.68  & 3.16  & 3.23  & 3.22  & 2.83  & 3.09  \\
\bottomrule
\end{tabular}\\[4pt]
\begin{tabular}{lcccccc}
\toprule
$K$ & 1 & 2 & 3 & 5$^{\dagger}$ & 10 & 25 \\
\midrule
Avg Acc ($\uparrow$)      & 89.99 & 89.19 & 89.70 & 90.59 & 89.04 & 89.60 \\
Forgetting ($\downarrow$) & 3.17  & 3.77  & 3.21  & 2.68  & 4.99  & 3.49  \\
\bottomrule
\end{tabular}\\[4pt]
\begin{tabular}{lccccccccccc}
\toprule
$r$ & 1 & 2 & 4 & 6 & 8 & 10$^{\dagger}$ & 12 & 14 & 16 & 18 & 20 \\
\midrule
Avg Acc ($\uparrow$)      & 90.42 & 90.39 & 90.25 & 90.30 & 90.40 & 90.59 & 90.29 & 90.25 & 90.20 & 90.10 & 90.08 \\
Forgetting ($\downarrow$) & 2.78  & 2.84  & 3.05  & 3.00  & 2.96  & 2.68  & 3.00  & 3.02  & 3.11  & 3.09  & 3.07  \\
\bottomrule
\end{tabular}\\[4pt]
\begin{tabular}{lccccccccccc}
\toprule
$\lambda$  & 0.1 & 1 & 5 & 10 & 50 & 60$^{\dagger}$ & 70 & 80 & 100 & 150 & 200 \\
\midrule
Avg Acc ($\uparrow$)      & 89.65 & 90.30 & 90.22 & 90.25 & 90.45 & 90.59 & 90.13 & 90.14 & 90.31 & 90.35 & 90.25 \\
Forgetting ($\downarrow$)  & 3.22  & 2.96  & 3.03  & 3.39  & 2.85  & 2.68  & 3.13  & 3.09  & 2.69  & 2.86  & 2.81  \\
\bottomrule
\end{tabular}
\end{table*}

\section{Additional analysis}
\label{supp:analysis}

\subsection{Analysis of the token evaluation module}
\label{supp:token_eval}

\paragraph{Sparse selection versus continuous weighting.}
Table~\ref{tab:routing_comparison} compares token weighting mechanisms
across the three benchmarks, isolating the weighting scheme on top of
the shared-prompt baseline.

\begin{table*}[t]
\centering
\caption{Performance comparison between soft token weighting and sparse selection. TAP is turned off during this experiment.}
\label{tab:routing_comparison}
\small
\setlength{\tabcolsep}{4pt}
\renewcommand{\arraystretch}{1.1}
\begin{tabular}{lcccccc}
\toprule
& \multicolumn{2}{c}{\textbf{CIFAR-100}} & \multicolumn{2}{c}{\textbf{ImageNet-R}} & \multicolumn{2}{c}{\textbf{CUB-200}} \\
\cmidrule(lr){2-3} \cmidrule(lr){4-5} \cmidrule(lr){6-7}
\textbf{Token weighting strategy} & \textbf{Avg Acc} ($\uparrow$) & \textbf{Forgetting} ($\downarrow$) & \textbf{Avg Acc} ($\uparrow$) & \textbf{Forgetting} ($\downarrow$) & \textbf{Avg Acc} ($\uparrow$) & \textbf{Forgetting} ($\downarrow$) \\
\midrule
Sparse selection (Top-$K$) & 84.53 $\pm$ 0.27         & {5.58 $\pm$ 0.33}        & 75.01 $\pm$ 0.30         & 6.39 $\pm$ 0.21         & 68.87 $\pm$ 0.42         & 11.62 $\pm$ 0.61        \\
Softmax weighting &      85.16 $\pm$ 0.42    &   5.91 $\pm$ 0.35    &    75.24 $\pm$ 0.33     &  5.36 $\pm$ 0.27        & 69.43 $\pm$ 0.43         & 11.75 $\pm$ 0.34  \\
Sigmoid weighting  & \textbf{86.73 $\pm$ 0.45} & \textbf{5.32 $\pm$ 0.36} & \textbf{75.46 $\pm$ 0.32} & \textbf{5.21 $\pm$ 0.35} & \textbf{69.59 $\pm$ 0.54} & \textbf{11.52 $\pm$ 0.42} \\
\bottomrule
\end{tabular}
\end{table*}

\begin{table}[t]
\centering
\caption{Variable on which the anchor statistics are computed
(CIFAR-100, 10 tasks, single seed).}
\label{tab:anchor_stats_variable}
\small
\setlength{\tabcolsep}{4pt}
\renewcommand{\arraystretch}{1.1}
\begin{tabular}{lcc}
\toprule
Anchor statistics over &  $\bar{\mathbf{q}}$  & $\mathbf{Q}$ \\
\midrule
Avg Acc ($\uparrow$)      & 90.59 & 90.02 \\
Forgetting ($\downarrow$) & 2.68  & 3.01  \\
Statistics memory (MB)    & 3.7   & 3.7   \\
\bottomrule
\end{tabular}
\end{table}

\begin{itemize}
    \item Continuous weighting (Softmax or Sigmoid) consistently enhances
    accuracy across all datasets when compared to sparse selection,
    while maintaining effective control over forgetting. This
    improvement arises from the ability of soft weighting to manage
    token selection more precisely.

    \item When comparing Sigmoid and Softmax weighting, Sigmoid
    consistently performs better on both accuracy and forgetting
    across all benchmarks. This aligns with the observation that softmax gating can cause “unnecessary competition among experts” and lead to representation collapse \cite{nguyen2024sigmoid}. Therefore, sigmoid gating prevents this drawback and enables more effective collaboration between tokens.
\end{itemize}

\begin{figure*}[t]
\centering
\includegraphics[width=\textwidth]{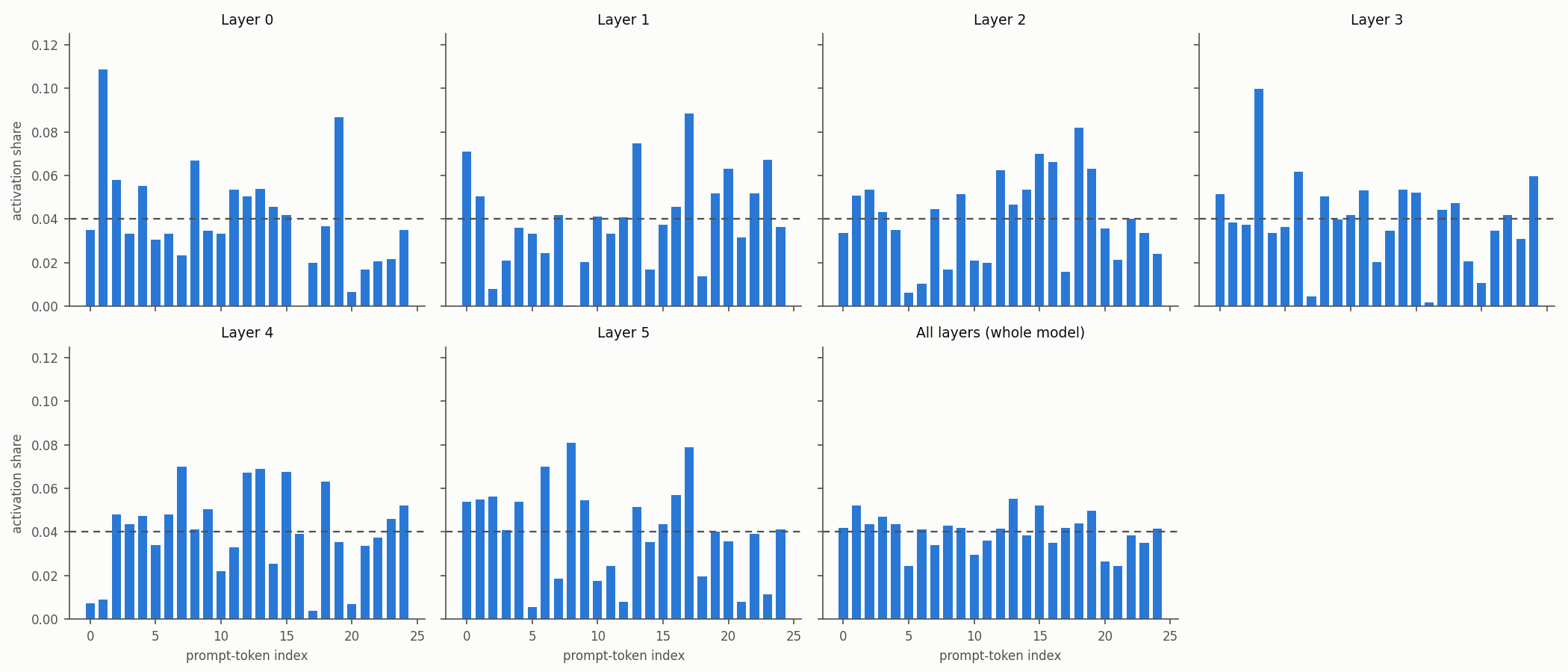}
\caption{Prompt-token utilization pooled over all 10 tasks and all 12
heads of each injected layer (last panel: all layers), the
dashed line marks the uniform level $1/25$.}
\label{fig:utilization}
\end{figure*}

\begin{figure*}[t]
\centering
\includegraphics[width=0.9\textwidth]{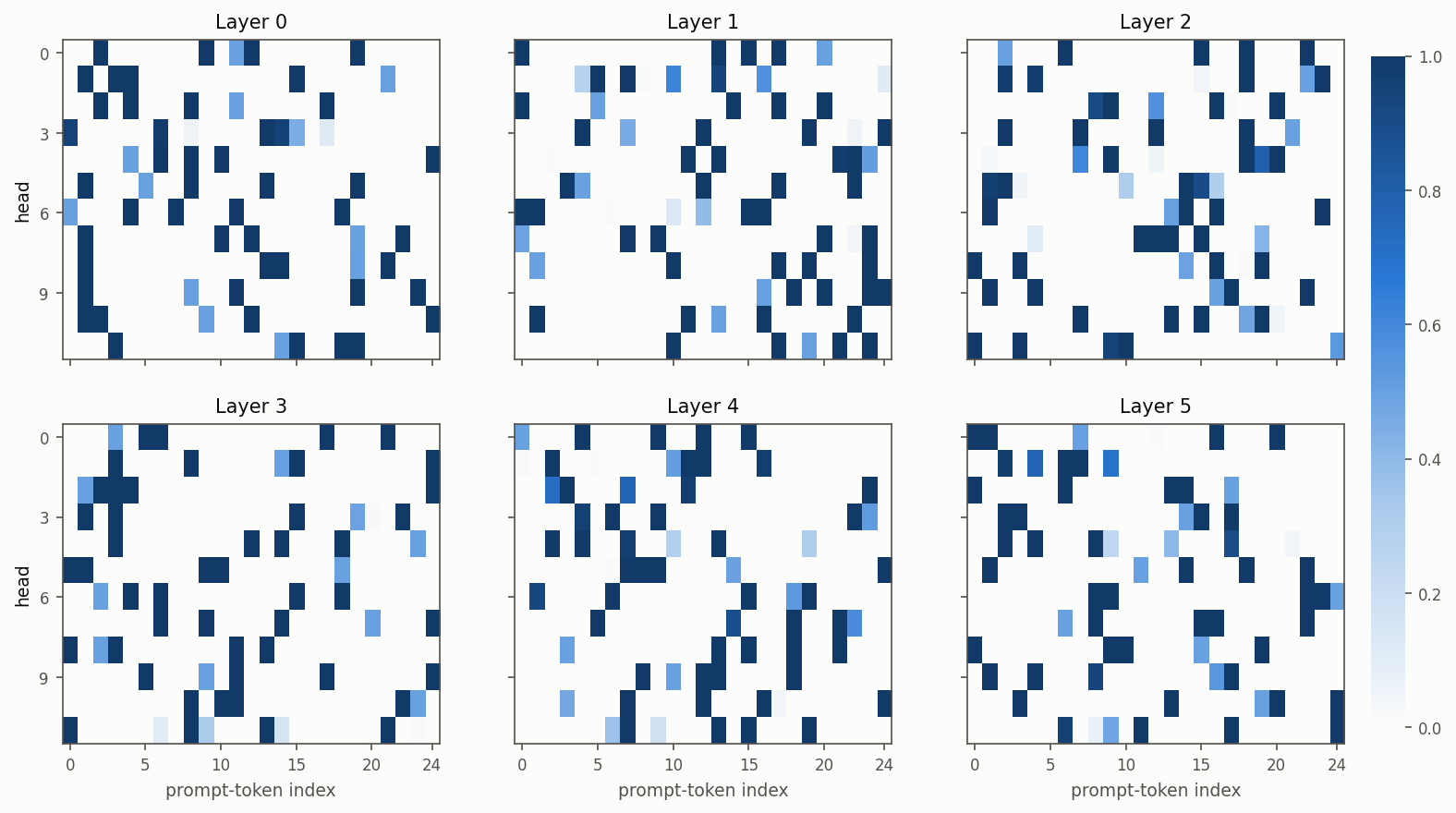}
\caption{Mean evaluation weight $\gamma$ on Task 2, per head at each of the six injected layers; each cell averages the
weight over all Task-2 training images.}

\label{fig:gate_heatmap}
\end{figure*}

\paragraph{Utilization and specialization of the prompt tokens.}
There are two natural concerns regarding the continuous weighting mechanism: either it causes a small number of tokens to dominate (leading to collapse), or it maintains partial activation for all tokens across all inputs, resulting in cross-task interference.
Figures~\ref{fig:utilization} and~\ref{fig:gate_heatmap} demonstrate that TEA-Prompt avoids these two issues.
When aggregating results across all tasks and attention heads, every prompt token receives an above-threshold activation rate that is remarkably uniform (Figure~\ref{fig:utilization}), without suffering from load-balancing issues. However, within the scope of a specific task, the evaluation weights concentrate attention on a select few tokens while reducing attention on the others
(Figure~\ref{fig:gate_heatmap}). Thus, we can conclude that there is stability in token activation alongside specialization as tokens adapt to specific inputs, thereby mitigating cross-task interference.

\paragraph{Are low-weight tokens informative?}
Table~\ref{tab:gate_dist} quantifies the near-binary behavior of
Fig.~\ref{fig:gate_heatmap}: $70.3\%$ of all evaluation weights lie below $0.1$ and $20\%$ above $0.5$, leaving only $\sim$$10\%$ in the transition band $[0.1,0.5)$. Whether this retained low-weight matters is tested in Table~\ref{tab:masking}, which discards all prompt
tokens whose weight falls below an increasing threshold. Removing the
$70\%$ of tokens with $\gamma<0.1$ already costs $0.6$ points of
accuracy and $0.2$ of forgetting; removing everything below $0.5$ costs $1.5$ points of accuracy and $0.6$ of forgetting, consistent with the gap between sparse selection
and sigmoid weighting in Table~\ref{tab:routing_comparison}. Thus, while low-weight tokens contribute little individually, they carry useful information in the aggregate due to their sheer volume. When multiplied by the squared difference in attention values between the current and previous tasks, they can significantly influence the distillation loss $\mathcal{L}_{\mathrm{AD}}$. Consequently, retaining this "long tail" enhances the plasticity of TEA-Prompt while maintaining its effectiveness in mitigating forgetting.

\begin{table}[t]
\centering
\caption{Distribution of the evaluation weights $\gamma$ over all
(input, head, token) activations on CIFAR-100. The $20\%$ mass above
$0.5$ equals $K/M$ by construction of the threshold.}
\label{tab:gate_dist}
\small
\setlength{\tabcolsep}{5pt}
\renewcommand{\arraystretch}{1.1}
\begin{tabular}{lcc}
\toprule
$\gamma$ interval & Share (\%) & Cumulative (\%) \\
\midrule
$[0,\,0.1)$   & 70.28 & 70.28  \\
$[0.1,\,0.2)$ & 6.60  & 76.88  \\
$[0.2,\,0.3)$ & 1.81  & 78.69  \\
$[0.3,\,0.4)$ & 0.79  & 79.48  \\
$[0.4,\,0.5)$ & 0.52  & 80.00  \\
$[0.5,\,1]$   & 20.00 & 100.00 \\
\bottomrule
\end{tabular}
\end{table}

\begin{table}[t]
\centering
\caption{Removing low-weight prompt tokens on CIFAR-100 (10 tasks).
Tokens whose evaluation weight falls below $\mathrm{thr}$ are discarded
(weight set to zero) throughout training and inference;
$\mathrm{thr}{=}0$ is the unmodified model. The second row gives the
fraction of activations removed (from Table~\ref{tab:gate_dist}).}
\label{tab:masking}
\small
\setlength{\tabcolsep}{3pt}
\renewcommand{\arraystretch}{1.1}
\begin{tabular}{lcccccc}
\toprule
$\mathrm{thr}$            & 0 (none) & 0.1   & 0.2   & 0.3   & 0.4   & 0.5   \\
\midrule
Removed (\%)              & 0        & 70.28 & 76.88 & 78.69 & 79.48 & 80.00 \\
Avg Acc ($\uparrow$)      & 90.59    & 89.95 & 89.75 & 90.25 & 89.52 & 89.10 \\
Forgetting ($\downarrow$) & 2.68     & 2.87  & 3.14  & 2.96  & 3.20  & 3.24  \\
\bottomrule
\end{tabular}
\end{table}

\subsection{Analysis of the anchoring loss}
\label{supp:anchor}

\begin{table}[htbp]
\centering
\caption{Comparison of task-specific query and class-specific query for $\mathcal{L}_{\mathrm{AD}}$ on CIFAR-100.}
\label{tab: class query}
\footnotesize %
\setlength{\tabcolsep}{4pt} %
\resizebox{\linewidth}{!}{%
\begin{tabular}{lcc}
\toprule
\textbf{\methodname{}} & \textbf{Avg Acc} ($\uparrow$) & \textbf{Forgetting} ($\downarrow$) \\
\midrule
 without $\mathcal{L}_{\mathrm{AD}}$ & 89.25 $\pm$ 0.21 & 3.68 $\pm$ 0.38 \\
with $\mathcal{L}_{\mathrm{AD}}$ (task-specific query) & 89.74 $\pm$ 0.25 & 3.57 $\pm$ 0.29\\
with $\mathcal{L}_{\mathrm{AD}}$ (class-specific query) & \textbf{90.59} $\pm$ 0.31 & \textbf{2.68} $\pm$ 0.25 \\
\bottomrule
\end{tabular}%
}
\end{table}

\paragraph{Why an instance-level query?}
The reduced query $\bar{\mathbf{q}}$ plays a dual role in \methodname{}:
it is the input of the token evaluation module, and it is the variable whose class-conditional Gaussian statistics drive the anchoring loss . For the first role, instance-level evaluation is equivalent to the prompt-attention score aggregation of SMoPE~\cite{le2025one},
which provably preserves the sample complexity of prompt estimation and empirically matches the convergence of
standard prefix tuning. We therefore do not re-evaluate that design
choice and focus on the impact of it on our anchoring loss $\mathcal{L}_{\mathrm{AD}}$. Table \ref{tab:anchor_stats_variable} compares the variations observed when using the reduced query $\bar{\mathbf{q}}$ versus the raw query $\mathbf{Q}$. Due to linearity, both variants share the same class cluster centers; they differ only in terms of dispersion, even memory is the same. This "dilution" reduces stability, thereby impairing memorization capability. Thus, $\bar{\mathbf{q}}$ enhances the statistical quality of the data "anchor," rather than merely offering convenience during the collection process.

\paragraph{Class-specific versus task-specific queries.}
A further comparison between task-specific and class-specific queries
for $\mathcal{L}_{\mathrm{AD}}$ is reported in
Table~\ref{tab: class query}, verifying the formulation of the
proposed anchoring loss. As shown in the table, there is a clear gap
between the two: task-specific queries provide only coarse-grained,
aggregated guidance across the diverse sub-distributions within a
single task, whereas class-specific queries enforce more fine-grained
attention behavior, enabling the tokens to precisely preserve
class-level representations. These observations validate that the
anchoring loss with class-specific queries acts as a robust
regularizer, effectively minimizing inter-class representation overlap
and boosting the model's resistance to catastrophic forgetting.

\paragraph{Choice of the sampling density.}
The anchoring loss is a Monte-Carlo objective over queries sampled
from the stored class statistics; Table~\ref{tab:anchor_density}
ablates the density model used for this sampling.

\begin{table}[t]
\centering
\caption{Density model used to sample anchor queries
$\tilde{\mathbf{q}}$ in $\mathcal{L}_{\mathrm{AD}}$ (CIFAR-100,
10 tasks, single seed). GFLOPs measure one anchor
evaluation per training step at the final task. $^{\dagger}$: diagonal
Gaussian (ours); Mean only = $\sigma{=}0$; Full covariance Gaussian; GMM-2 with $2$ diagonal Gaussian components. TAP is turn-off during this experiment.}
\label{tab:anchor_density}
\small
\setlength{\tabcolsep}{4pt}
\renewcommand{\arraystretch}{1.1}
\begin{tabular}{lcccc}
\toprule
Sampling model & Diag.$^{\dagger}$ & Mean only & Full-cov. & GMM-2 \\
\midrule
Avg Acc ($\uparrow$)      & 87.37 & 87.02 & 87.25 & 86.15 \\
Forgetting ($\downarrow$) & 3.96  & 4.09  & 4.08  & 3.98  \\
GFLOPs/step               & 2.19  & 0.22  & 2.72  & 2.20  \\
\bottomrule
\end{tabular}
\end{table}

More complex density models offer no benefit: a Gaussian distribution with a full covariance matrix and a two-component diagonal Gaussian Mixture Model (GMM-2) yield accuracy comparable to a diagonal Gaussian without improving resistance to forgetting, while incurring up to 24\% higher computational costs per step (2.72 GFLOPS vs 2.19 GFLOPS). Completely removing the variance component (retaining only the mean) is the least expensive option but compromises accuracy. Using a single fixed probe per class fails to cover the entire query region that needs to be preserved. This confirms that the distribution of samples holds the signal for anchoring information rather than just the class center. Consequently, the diagonal Gaussian achieves an optimal balance between accuracy and cost. Nevertheless, this total cost remains a small fraction of the forward pass for a single image, enabling TEA-Prompt to achieve strong performance in the single shared prompt scenario.

\end{document}